\documentclass[10pt,twocolumn,letterpaper]{article}
\usepackage{cvpr}
\usepackage{tikz}
\usetikzlibrary{positioning,calc,matrix,arrows}
\usepackage{pgfplots}
\usepackage{times}
\usepackage{graphicx}
\usepackage{amsmath}
\usepackage{amssymb}
\usepackage{booktabs}
\usepackage{array}
\usepackage{mathtools}
\usepackage[breaklinks=true,bookmarks=false]{hyperref}
\usepackage{cleveref}

\tikzstyle{block} = [draw, rectangle, minimum height=3em, minimum width=3em]
\tikzstyle{data} = []
\tikzstyle{datac} = [draw, circle, minimum height=1em, minimum width=1em,inner sep=3pt]
\tikzstyle{par} = [draw, circle, minimum height=1em, minimum width=1em,fill=black!20,inner sep=3pt]
\tikzstyle{pinstyle} = [pin edge={to-,thin,black}]
\tikzstyle{to} = [->,>=stealth',shorten >=1pt,semithick]
\tikzstyle{btwn} = [-,>=stealth',semithick]

\newcolumntype{L}[1]{>{\raggedright\let\newline\\\arraybackslash\hspace{0pt}}m{#1}}
\newcolumntype{R}[1]{>{\raggedleft\let\newline\\\arraybackslash\hspace{0pt}}m{#1}}
\newcolumntype{C}[1]{>{\centering\let\newline\\\arraybackslash\hspace{0pt}}m{#1}}

\newcolumntype{x}{>\small c}
\newcommand{\bx}{\mathbf{x}}
\newcommand{\by}{\mathbf{y}}
\newcommand{\bw}{\mathbf{w}}

\pdfoutput=1
\cvprfinalcopy 

\def\cvprPaperID{848} 

\ifcvprfinal\fi
\begin{document}

\title{Weakly Supervised Deep Detection Networks}
\author{Hakan Bilen\\
University of Oxford\\
{\tt\small hbilen@robots.ox.ac.uk}
\and
Andrea Vedaldi\\
University of Oxford\\
{\tt\small vedaldi@robots.ox.ac.uk}
}

\maketitle
\begin{abstract}
Weakly supervised learning of object detection is an important problem in image understanding that still does not have a satisfactory solution.  In this paper, we address this problem by exploiting the power of deep convolutional neural networks pre-trained on large-scale image-level classification tasks. We propose a weakly supervised deep detection architecture that modifies one such network to operate at the level of image regions, performing simultaneously region selection and classification. Trained as an image classifier, the architecture implicitly learns object detectors that are better than alternative weakly supervised detection systems on the PASCAL VOC data. The model, which is a simple and elegant end-to-end architecture, outperforms standard data augmentation and fine-tuning techniques for the task of image-level classification as well.

\end{abstract}

\section{Introduction}\label{s:intro}
In recent years, Convolutional Neural Networks (CNN) \cite{Lecun89} have emerged as the new state-of-the-art learning framework for image recognition. Key to their success is the ability to learn from large quantities of labelled data the complex appearance of real-world objects. One of the most striking aspects of CNNs is their ability to learn generic visual features that generalise to many tasks. In particular, CNNs pre-trained on datasets such as ImageNet ILSVRC have been shown to obtain excellent results in recognition in other domains~\cite{Donahue13}, in object detection \cite{Girshick14}, in semantic segmentation \cite{Hariharan14}, in human pose estimation \cite{Toshev13}, and in many other tasks.

In this paper we look at how the power of CNNs can be leveraged in  \emph{weakly supervised detection} (WSD), which is the problem of learning object detectors using only image-level labels. The ability of learning from weak annotations is very important for two reasons: first, image understanding aims at learning an growing body of complex visual concepts (\eg hundred thousands object categories in ImageNet). Second, CNN training is data-hungry. Therefore, being able to learn complex concepts using only light supervision can reduce significantly the cost of data annotation in tasks such as image segmentation, image captioning, or object detection.

\begin{figure}
\begin{center}
\includegraphics[width=\columnwidth]{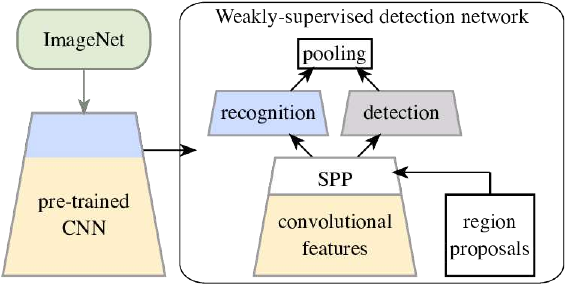} 
\end{center}
\caption{{\bf Weakly Supervised Deep Detection Network.} Our method starts from a CNN pre-trained for image classification on a large dataset, e.g. ImageNet. It then modifies to reason efficiently about regions, branching off a recognition and a detection data streams. The resulting architecture can be fine-tuned on a target dataset to achieve state-of-the-art weakly supervised object detection using only image-level annotations.}
\label{f:splash}
\end{figure}

We are motivated in our research by the hypothesis that, since pre-trained CNNs generalise so well to a large number of tasks, they should contain meaningful representations of the data. For example, there exists evidence that CNNs trained for image classification learn proxies to objects and objects parts~\cite{Zhou15}. Remarkably, these concepts are acquired implicitly, without ever providing the network with information about the \emph{location} of such structures in images. Hence, CNNs trained for image classification may already contain implicitly most of the information required to perform object detection.

We are not the first to address the problem of WSD with CNNs. The method of Wang~\etal~\cite{Wang14a}, for example, uses a pre-trained CNN to describe image regions and then learn object categories as corresponding visual topics. While this method is currently state-of-the-art in weakly supervised object detection, it comprises several components beyond the CNN and requires signifiant tuning. 

In this paper we contribute a novel {\em end-to-end} method for weakly supervised object detection using pre-trained CNNs which we call a \emph{weakly supervised deep detection network} (WSDDN) (\cref{f:splash}). Our method (\cref{s:method}) starts from an existing network, such as AlexNet pre-trained on ImageNet data, and extends it to reason explicitly and efficiently about image regions $R$. In order to do so, given an image $\bx$, the first step is to efficiently extract region-level descriptors $\phi(\bx;R)$ by inserting a spatial pyramid pooling layer on top of the convolutional layers of the CNN~\cite{He14,Girshick15}. Next,  the network is branched to extract \emph{two data streams} from the pooled region-level features. The first stream associates a class score $\phi^c(\bx;R)$ to each region individually, performing \emph{recognition}.  The second stream, instead, \emph{compares} regions by computing a probability distribution $\phi^d(\bx;R)$ over them; the latter represents the belief that, among all the candidate regions in the image,  $R$ is the one that contains the most salient image structure, and is therefore a proxy to \emph{detection}. The recognition and detection scores computed for all the image regions are finally aggregated in order to predict the class of the image as a whole, which is then used to inject image-level supervision in learning.

It is interesting to compare our method to the most common weakly supervised object detection technique, namely multiple instance learning (MIL)~\cite{Dietterich97}. MIL alternates between selecting which regions in images look like  the object of interest and estimating an appearance model of the object using the selected regions. Hence, MIL uses the appearance model itself to perform region selection. Our technique differs from MIL in a fundamental way as regions are selected by a dedicated \emph{parallel detection branch} in the network, which is independent of the recognition branch. In this manner, our approach helps avoiding one of the pitfalls of MIL, namely the tendency of the method to get stuck in local optima. 

Our two-stream CNN is also weakly related to the recent work of Lin~\etal~\cite{Lin15}. They propose a ``bilinear'' architecture where the output of two parallel network streams are combined by taking the outer product of feature vectors at corresponding spatial locations. The authors state that this construction is inspired by the ventral and dorsal streams of the human visual system, one focusing on recognition and the other one on localisation. While our architecture contains two such streams, the similarity is only superficial. A key difference is that in Lin~\etal the two streams are perfectly symmetric, and therefore there is no reason to believe that one should perform classification and the other detection; in our scheme, instead, the detection branch is explicitly designed to compare regions, breaking the symmetry. Note also that Lin~\etal~\cite{Lin15} do not perform WSD nor evaluate object detection performance.

Once the modifications have been applied, the network is ready to be fine-tuned on a target dataset, using only image-level labels, region proposals and back-propagation. In \cref{s:experiments} we show that, when fine-tuned on the PASCAL VOC training set, this architecture achieves state-of-the-art weakly supervised object detection on the PASCAL data, achieving superior results to the current state-of-the-art~\cite{Wang14a} but \emph{using only CNN machinery}. Since the system can be trained end-to-end using standard CNN packages, it is also as efficient as the recent fully-supervised Fast R-CNN detector of Girshick~\etal~\cite{Girshick15}, both in training and in testing. Finally, as a byproduct of our construction we also obtain a powerful image classifier that \emph{performs better than standard fine-tuning techniques} on the target data. Our findings are summarised in~\cref{s:conclusions}.

\section{Related Work}\label{s:related}
The majority of existing approaches to WSD formulate this task as  MIL. In this formulation an image is interpreted as a bag of regions. If the image is labeled as positive, then one of the regions is assume to tightly contain the object of interest. If the image is labeled as negative, then no region contains the object. Learning alternates between estimating a model of the object appearance and selecting which regions in the positive bags correspond to the object using the appearance model. 

The MIL strategy results in a non-convex optimization problem; in practice, solvers tend to get stuck in local optima such that the quality of the solution strongly depends on the initialization. Several papers have focused on developing various initialization strategies \cite{Kumar10a,Deselaers10,Song14a,Cinbis15} and on regularizing the optimization problem \cite{Song14,Bilen14}. Kumar~\etal \cite{Kumar10a} propose a self-paced learning strategy that progressively includes harder samples to a small set of initial ones at training. Deselaers~\etal~\cite{Deselaers10} initialize object locations based on the objectness score. Cinbis~\etal \cite{Cinbis15} propose a multi-fold split of the training data to escape local optima. Song~\etal~\cite{Song14} apply Nesterov's smoothing technique \cite{Nesterov05} to the latent SVM formulation \cite{Felzenszwalb10a} to be more robust against poor initializations. Bilen~\etal~\cite{Bilen14} propose a smoothed version of MIL that softly labels object instances instead of choosing the highest scoring ones. Additionally, their method regularizes the latent object locations by penalizing unlikely configurations based on symmetry and mutual exclusion principles.

Another line of research in WSD \cite{Song14,Song14a,Wang14a} is based on the idea of identifying the similarity between image parts. Song~\etal \cite{Song14} propose a discriminative graph-based algorithm that selects a subset of windows such that each window is connected to its nearest neighbors in positive images. In~\cite{Song14a}, the same authors extend this method to discover multiple co-occurring part configurations. Wang~\etal~\cite{Wang14a} propose an iterative technique that applies a latent semantic clustering via latent Semantic Analysis (pLSA) on the windows of positive samples and selects the most discriminative cluster for each class based on its classification performance. Bilen~\etal \cite{Bilen15} propose a formulation that jointly learns a discriminative model and enforces the similarity of the selected object regions via a discriminative convex clustering algorithm.

Recently a number of researchers \cite{Oquab14,Oquab15} have proposed weakly supervised localization principles to improve classification performance of CNNs without providing any annotation for the location of objects in images. Oquab~\etal \cite{Oquab14} employ a pre-trained CNN to compute a mid-level image representation for images of PASCAL VOC. In their follow-up work, Oquab~\etal \cite{Oquab15} modify a CNN architecture to \emph{coarsely} localize object instances in image while predicting its label. 

Jaderberg~\etal~\cite{Jaderberg15c} proposed a CNN architecture in which a subnetwork automatically pre-transforms an image in order to optimize the classification accuracy of a second subnetwork. This ``transformer network'', which is trained in an end-to-end fashion from image-level labels, is shown to align objects to a common reference frame, which is a proxy to detection. Our architecture contains a mechanism that pre-select image regions that are likely to contain the object, also trained in an end-to-end fashion; while this may seem very different, this mechanism can also be thought as learning transformations (as the ones that map the detected regions to a canonical reference frame). However, the nature of the selection process in in our and their networks are very different.

\section{Method}\label{s:method}
 \begin{figure*}[th]
\begin{center}
\footnotesize
\begin{tikzpicture}[auto, node distance=0.5cm]
\node (x) {$\bx$} ;
\node (xsplit) [right=of x] {};
\matrix (m) [matrix of math nodes, 
    column sep=0.5cm,
    row sep=0.2cm,
    right=of xsplit]
 { 
\node(pool5) [block] {\phi_\text{pool5}} ;
& \node(spp) [block] {\phi_\text{SPP}} ;
& \node(fc6) [block] {\phi_\text{fc6}};
& \node(fc7) [block] {\phi_\text{fc7}};
& \node(fc8) [block] {\phi_\text{fc8c}};
\\
\node(ssw) [block] {\text{SSW/EB}};
& &&& \node(fc8d) [block] {\phi_\text{fc8d}};
\\
};
\node [block,right=1cm of fc8] (cscore) {$\sigma_\text{class}$};
\node [block,right=1cm of fc8d] (dscore) {$\sigma_\text{det}$};
\node [draw, rectangle,below right=-0.05cm and 0.5cm of cscore] (times) {$\odot$};
\node [draw, rectangle,right=of times] (sum) {$\Sigma$};
\node [right=of sum] (y) {$\by$};
\draw[to] (x) -- (0.75,0) |- (pool5);
\draw[to] (pool5) -- (spp) ;
\draw[to] (spp) -- (fc6) ;
\draw[to] (fc6) -- (fc7) ;
\draw[to] (fc7) -- (fc8) ;
\draw[to] (fc8) -- node {$\bx^c$} (cscore) ;
\draw[to] (0.75,0) |- (ssw) ;
\draw[to] (ssw) -| node[right] {$\mathcal{R}$} (spp) ; 
\draw[to] (fc7) |- (fc8d);
\draw[to] (fc8d) -- node {$\bx^d$} (dscore) ;
\draw[to] (cscore) -| (times) ;
\draw[to] (dscore) -| (times) ;
\draw[to] (times) -- node {$\bx^{\mathcal{R}}$} (sum) ;
\draw[to] (sum) -- (y) ;
\end{tikzpicture}
\end{center}
\label{f:WSDDN}
\caption{{\bf Weakly-supervised deep detection network.} The figure illustrates the architecture of WSDDN.}
\end{figure*}
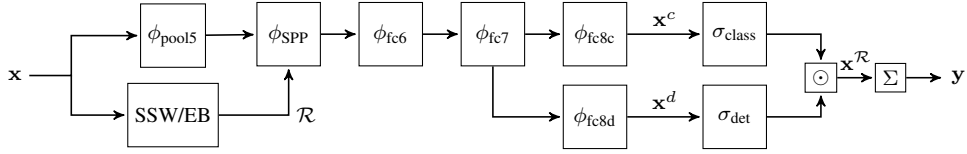

In this section we introduce our \emph{weakly supervised deep detection network} (WSDDN) method. The overall idea consists of three steps. First, we obtain a CNN pre-trained on a large-scale image classification task (\cref{s:pretrained}). Second, we construct the WSDDN as an architectural modification of this CNN (\cref{s:WSDDN}). Third, we train/fine-tune the WSDDN on a target dataset, once more using only image-level annotations (\cref{s:training}). The remainder of this section discusses these three steps in detail.
 
\subsection{Pre-trained network}\label{s:pretrained}
We build our method on a pre-trained CNN that has been pre-trained on the ImageNet ILSVRC 2012 data \cite{Russakovsky15} with only image-level supervision (\ie no bounding box annotations). We give the details of the used CNN architectures in \cref{s:experiments}. 
\subsection{Weakly supervised deep detection network}\label{s:WSDDN}

Given the pre-trained CNN, we transform it into a WSDDN by introducing three modifications (see also \cref{f:WSDDN}). First, we replace the last pooling layer immediately following the ReLU layer in the last convolutional block (also known as \textit{relu5} and \textit{pool5}, respectively) with a layer implementing \emph{spatial pyramid pooling} (SPP)~\cite{Lazebnik06,He14}. This results in a function that takes as input an image $\bx$ and a region (bounding box) $R$ and produces as output a feature vector or representation $\phi(\bx;R)$. Importantly, the function decomposes as
\[
\phi(\bx;R) = \phi_{\text{SPP}}(\cdot ; R) \circ \phi_{\text{relu5}}(\bx)
\]
where $\phi_{\text{relu5}}(\bx)$ needs to be computed only once for the whole image and $\phi_{\text{SPP}}(\cdot ; R)$ is fast to compute for any given region $R$. In practice, SPP is configured to be compatible to the first fully connected layers of networks (\ie fc6). Note that SPP is implemented as a network layer as in~\cite{Girshick15} to allow to train the system end-to-end (and for efficiency).

Given an image $\bx$, a shortlist of candidate object regions $\mathcal{R} = (R_1,\dots,R_n)$ are obtained by a region proposal mechanism. Here we experiment with two methods, Selective Search Windows (SSW)~\cite{Sande11} and Edge Boxes (EB)~\cite{Zitnick14}. As in~\cite{Girshick15}, we then modify the SPP layer to take as input not a single region, but rather the full list $\mathcal{R}$; in particular, $\phi(\bx;\mathcal{R})$ is defined as the concatenation of $\phi(\bx;R_1), \dots, \phi(\bx;R_n)$ along the fourth dimension (since each individual $\phi(\bx;R)$ is a 3D tensor).

At this point in the architecture, region-level features are further processed by two fully connected layers $\phi_\text{fc6}$ and $\phi_\text{fc7}$, each comprising a linear map followed by a ReLU. Out of the output of the last such layer, we branch off two data streams, described next.

\paragraph{Classification data stream.} The first data stream performs \emph{classification} of the individual regions, by mapping each of them to a $C$-dimensional vector of class scores, assuming that the system is trained to detect $C$ different classes. This is achieved by evaluating a linear map $\phi_{fc8c}$ and results in a matrix of data $\bx^c \in \mathbb{R}^{C \times |\mathcal{R}|}$, containing the class prediction scores for each region. The latter is then passed through a \emph{softmax} operator, defined as follows:
\begin{equation}\label{e:softmaxc}
  [\sigma_\text{class}(\bx^c)]_{ij}
  = 
  \frac{e^{x^c_{ij}}}{\sum_{k=1}^{C} e^{x^c_{kj}}}.
\end{equation}

\paragraph{Detection data stream.} The second data stream performs instead \emph{detection}, by scoring regions relative to one another. This is done on a class-specific basis by using a second linear map $\phi_{fc8d}$, also resulting in a matrix of scores $\bx^d \in \mathbb{R}^{C \times |\mathcal{R}|}$. It is then passed through another softmax operator, but this time defined as follows:
\begin{equation}\label{e:softmaxd}
  [\sigma_\text{det}(\bx^d)]_{ij}
  = 
  \frac{e^{x^d_{ij}}}{\sum_{k=1}^{|\mathcal{R}|} e^{x^d_{ik}}}.
\end{equation}

While the two streams are remarkably similar, the introduction of the $\sigma_\text{class}$ and $\sigma_\text{det}$ non-linearities in the classification and detection streams is a key difference which allows to interpret them as performing classification and detection, respectively. In the first case, in fact, the softmax operator compares, for each region independently, class scores, whereas in the second case the softmax operator compares, for each class independently, the scores of different regions. Hence, the first branch predicts which class to associate to a region, whereas the second branch selects which regions are more likely to contain an informative image fragment.

\paragraph{Combined region scores and detection.} The final score of each region is obtained by taking the element-wise (Hadamard) product $\bx^\mathcal{R} = \sigma_\text{class}(\bx^c) \odot \sigma_\text{det}(\bx^d)$ of the two scoring matrices. The region scores are then used to rank image regions by likelihood of centring an object (for each class independently); standard non-maxima suppression is then performed (by iteratively removing regions with Intersection over Union (IoU) larger than 40\% with regions already selected) to obtain the final list of class-specific detections in an image.

The way the two streams' scores are combined is reminiscent of the bilinear networks of~\cite{Lin15}, but there are three key differences. The first difference is that the introduction of the different softmax operators explicitly breaks the symmetry of the two streams. The second one is that, instead of computing the outer product of the two feature vectors $\sigma_\text{class}(\bx^c_r) \otimes \sigma_\text{det}(\bx^d_r)$, we compute the element-wise product $\sigma_\text{class}(\bx^c_r) \odot \sigma_\text{det}(\bx^d_r)$ (generating quadratically less parameters). The third difference is that scores $\sigma_\text{class}(\bx^c_r) \otimes \sigma_\text{det}(\bx^d_r)$ are computed for specific image regions $r$ rather than a fixed set of image locations on a grid. Together, these three differences mean that we can interpret $\sigma_\text{det}(\bx^d)$ as a term that ranks regions, whereas $\sigma_\text{class}(\bx^c)$ ranks classes. It is more difficult to clearly assess the nature of the two streams in~\cite{Lin15}.

\paragraph{Image-level classification scores.} So far, WSDDN has computed region-level scores $\bx^\mathcal{R}$. This is transformed in an image-level class prediction score by summation over regions:
\[
   y_c = \sum_{r=1}^{|\mathcal{R}|} x^{\mathcal{R}}_{cr}.
\] 
Note that both $y_c$ is a sum of element-wise product of softmax normalised scores over $|\mathcal{R}|$ regions and thus it is in the range of $(0,1)$. Softmax is not performed at this stage as images are allowed to contain more than one object class (whereas regions should contain a single class).

\subsection{Training WSDDN}\label{s:training}

Having discussed the WSDDN architecture in the previous section, here we explain how the model is trained. The data is a collection of images $\bx_i, i=1,\dots, n$ with \emph{image level labels} $\by_i \in \{-1,1\}^C$. We denote by $\phi^{\by}(\bx|\bw)$ the complete architecture, mapping an image $\bx$ to a vector of class scores $\by\in\mathbb{R}^C$. The parameters $\bw$ of the model lump together the coefficients of all the filters and biases in the convolutional and fully-connected layers. Then, stochastic gradient descent with momentum is used to optimise the energy function
\begin{equation}
E(\bw)
=
\frac{\lambda}{2}\|\bw\|^2
+
\sum_{i=1}^n
\sum_{k=1}^C
\log(y_{ki} (\phi^{\by}_k(\bx_i|\bw)-\frac{1}{2}) + \frac{1}{2}),
\label{eq:energy}
\end{equation} hence optimising a sum of $C$ binary-log-loss terms, one per class. As $\phi^{\by}_k(\bx_i|\bw)$ is in range of $(0,1)$, it can be considered as a probability of class $k$ being present in image $\bx_i$, \ie $p(y_{ki}=1)$. When the ground-truth label is positive, the binary log loss becomes $\log(p(y_{ki}=1))$, $\log(1-p(y_{ki}=1))$ otherwise.

\subsection{Spatial Regulariser}
As WSDDN is optimised for image-level class labels, it does not guarantee any spatial smoothness such that if a region obtains a high score for an object class, the neighbouring regions with high overlap will also have high scores. In the supervised detection case, Fast-RCNN~\cite{Girshick15} takes the region proposals that have at least $50 \%$ IoU with a ground truth box as positive samples and learns to regress them into their corresponding ground truth bounding box. As our method does not have access to ground truth boxes, we follow a soft regularisation strategy that penalises the feature map discrepancies at the second fully connected layer~\texttt{fc7} between the highest scoring region and the regions with at least $60 \%$ IoU (\ie\; $r\in|\bar{R}|$) during training:
\[
\frac{1}{nC}
\sum_{k=1}^C
\sum_{i=1}^{N_{k}^+}
\sum_{r=1}^{|\bar{R}|}
\frac{1}{2}(\phi^{\by}_{k*i})^2(\phi^{\text{fc7}}_{k*i}-\phi^{\text{fc7}}_{kri})^{^\mathrm{T}}(\phi^{\text{fc7}}_{k*i}-\phi^{\text{fc7}}_{kri})
\] where $N_{k}^+$ is the number of positive images for the class $k$ and $*=\arg\max_r \phi^{\by}_{kri}$ is the highest scoring region in image $i$ for the class $k$. We add this regularisation term to the cost function in~\cref{eq:energy}.

\section{Experiments}\label{s:experiments}
In this section we conduct a thorough investigation of WSDDN and its components on weakly supervised detection and image classification.

\subsection{Benchmark data.}
We evaluate our method on the PASCAL VOC 2007 and 2010 datasets \cite{Everingham10}, as they are the most widely-used benchmark in weakly supervised object detection. While the VOC 2007 dataset consists of 2501 training, 2510 validation, and 5011 test images containing bounding box annotations for 20 object categories, VOC 2010 dataset contains 4998 training, 5105 validation, and 9637 test images for the same number of categories. We use the suggested training and validation splits and report results evaluated on \emph{test} split. We report performance of our method on both the object detection and the image classification tasks of PASCAL VOC. 

For detection, we use two performance measures. The first one follows the standard PASCAL VOC protocol and reports average precision (AP) at  $50\%$ intersection-over-union (IoU) of the detected boxes with the ground truth ones. We also report CorLoc, a commonly-used weakly supervised detection measure~\cite{Deselaers12}. CorLoc is the percentage of images that contain at least one instance of the target object class for which the most confident detected bounding box overlaps by at least $50\%$ with one of these instances. Differently from AP, which is measured on the PASCAL test set, CorLoc is evaluated on the union of the training and validation subset of PASCAL. For classification, we use the standard PASCAL VOC protocol and report AP.

\subsection{Experimental setup.}\label{subsec:expsetup}
We comprehensively evaluate our method with three pre-trained CNN models in our experiments as in~\cite{Girshick15}. The first network is the VGG-CNN-F~\cite{Chatfield14} which is similar to AlexNet~\cite{Krizhevsky12} but has reduced number of convolutional filters. We refer to this network as \textbf{S}, for small. The second one is VGG-CNN-M-1024 which has the same depth as \textbf{S} but has smaller stride in the first convolutional layer. We name this network \textbf{M} for medium. The last network is the deep VGG-VD16 model~\cite{Simonyan15} and we call this network \textbf{L} for large. These models, which are pre-trained on the ImageNet ILSVRC 2012 challenge data \cite{Russakovsky15}, attain  $18.8 \%$, $16.1 \%$ and $9.9 \%$ top-5 accuracy respectively (using a single centre-crop) on ILSVRC (importantly no bounding box information is provided during pre-training). As explained in \cref{s:pretrained}, we apply the following modifications to the network. First, we replace the last pooling layer \textit{pool5} with a SPP layer~\cite{He14} which is configured to be compatible with the network's first fully connected layer. Second, we add a parallel detection branch to the classification one that contains a fully-connected layer followed by a soft-max layer. Third, we combine the classification and detection streams by element-wise product followed by summing scores across regions, and feed the latter to a binary log-loss layer. Note that this layer assesses the classification performance for the 20 classes together, but each of them is treated as a different binary classification problem; the reason is that classes can co-occur in the PASCAL VOC, such that the softmax log loss used in AlexNet is not appropriate.

The WSDDNs are trained on the PASCAL VOC training and validation data by using fine-tuning on all layers, a widely-adopted technique to improve the performance of a CNN on a target domain~\cite{Chatfield14}. Here, however, fine tuning performs the essential function of learning the classification and detection streams, effectively causing the network to learn to detect objects, but using only weak image-level supervision. The experiments are run for $20$ epochs and all the layers are fine-tuned with the learning rate $10^{-5}$ for the first ten epochs and $10^{-6}$ for the last ten epochs. Each minibatch contains all region proposals from a single image.

In order to generate candidate regions to use with our networks, we evaluate two proposal methods, Selective Search Windows (SSW)~\cite{Sande11} using its \textit{fast} setting, and EdgeBoxes (EB)~\cite{Zitnick14}. In addition to region proposals, EB provides an objectness score for each region based on the number of contours wholly encloses. We exploit this additional information by multiplying the feature map $\phi_{\text{SPP}}$ proportional to its score via a scaling layer in WSDDN and denote this setting as \emph{Box Sc}. Since we use a SPP layer to aggregate descriptors for each region, images do not need to be resized to a particular size as in the original pre-trained model. Instead, we keep the original aspect ratio of images fixed and resize them to five different scales (setting their maximum of width or height to $\lbrace 480,576,688,864,1200\rbrace$ respectively) as in \cite{He14}. During training, we apply random horizontal flips to the images and select a scale at random as a form of jittering or data augmentation. At test time we average the outputs of 10 images (\ie the 5 scales and their flips). We use the publicly available CNN toolbox MatConvNet~\cite{Vedaldi15} to conduct our experiments and share our code, models and data~\footnote{\url{https://github.com/hbilen/WSDDN}}.

When evaluated on an image, WSDDN produces, for each target class $c$ and image $\bx$, a score $\bx^\mathcal{R}_r = S_c(\bx;r)$ for each region $r$ and an aggregated score $y_c=S_c(\bx)$ for each image. Non-maxima suppression (with $40$ \% IoU threshold) is applied to the regions and then the scored regions and images are pooled together to compute detection AP and CorLoc.

 \begin{table}
\begin{center}
\begin{tabular}{lccc|c}
& \textbf{S} & \textbf{M} & \textbf{L} & \textbf{Ens.}\\
\toprule
SSW                                       & 31.1 & 30.9 & 24.3 & 33.3 \\
EB                                        & 31.5 & 30.9 & 25.5 & 34.2 \\
EB + Box Sc. & 33.4 & 32.7 & 30.4 & 36.7\\
EB + Box Sc. + Sp. Reg.        & \bf{34.5} & \bf{34.9} & \bf{34.8} & \bf{39.3} \\
\end{tabular}
\vspace{0.2em}
\caption{{\bf VOC 2007 test} detection average precision (\%). The ensemble network is denoted as \textbf{Ens}.}
 \label{tab:voc2007base}
\end{center}
\end{table}

\begin{table*}[t!]
\centering
\renewcommand{\arraystretch}{1.2}
\renewcommand{\tabcolsep}{1.2mm}
\resizebox{\linewidth}{!}{
  \begin{tabular}{@{}L{3.5cm}|*{20}{x}|x@{}}
method & aero      & bike      & bird      & boat      & bottle     & bus        & car        & cat        & chair      & cow        & table      & dog        & horse      & mbike      & persn     & plant      & sheep      & sofa       & train      & tv         & mean       \\
\toprule
WSDDN \textbf{S} & 42.9 & 56.0 & 32.0 & 17.6 & 10.2 & 61.8 & 50.2 & 29.0 & 3.8 & 36.2 & 18.5 & 31.1 & \bf{45.8} & 54.5 & 10.2 & 15.4 & 36.3 & 45.2 & 50.1 & 43.8 & 34.5\\
WSDDN \textbf{M} & 43.6 & 50.4 & 32.2 & \bf{26.0} &  9.8 & 58.5 & 50.4 & 30.9 & 7.9 & 36.1 & 18.2 & 31.7 & 41.4 & 52.6 & 8.8 & 14.0 & 37.8 & 46.9 & 53.4 & 47.9 & 34.9\\ 
WSDDN \textbf{L} & 39.4 & 50.1 & 31.5 & 16.3 & 12.6 & 64.5 & 42.8 & \bf{42.6} & 10.1 & 35.7 & 24.9 & 38.2 & 34.4 & 55.6 & 9.4 & 14.7 & 30.2 & 40.7 & 54.7 & 46.9 & 34.8\\ 
WSDDN Ensemble   & 46.4 & \bf{58.3} & \bf{35.5} & {25.9} & \bf{14.0} & \bf{66.7} & \bf{53.0} & {39.2} & 8.9 & \bf{41.8} & \bf{26.6} & \bf{38.6} & {44.7} & \bf{59.0} & 10.8 & 17.3 & \bf{40.7} & \bf{49.6} & \bf{56.9} & \bf{50.8} & \bf{39.3} \\
\midrule
Bilen \etal \cite{Bilen14} & 42.2 & 43.9 & 23.1 &  9.2 & {12.5} & {44.9} & 45.1 & 24.9 &  8.3 & 24.0 & 13.9 & 18.6 & 31.6 & 43.6 &  7.6 & \bf{20.9} & 26.6 & 20.6 & 35.9 & 29.6 & 26.4\\ 
Bilen \etal \cite{Bilen15} & 46.2 & 46.9 & {24.1} & {16.4} & {12.2} & 42.2 & {47.1} & 35.2 &  7.8 & 28.3 & 12.7 & 21.5 & 30.1 & 42.4 &  7.8 & {20.0} & {26.8} & 20.8 & {35.8} & 29.6 & 27.7\\
Cinbis \etal \cite{Cinbis15} & 39.3 & 43.0 & 28.8 & 20.4 &  8.0 & 45.5 & 47.9 & 22.1 & 8.4 & 33.5 & 23.6 & 29.2 & 38.5 & 47.9 & \bf{20.3} & 20.0 & 35.8 & 30.8 & 41.0 & 20.1 & 30.2 \\
Wang \etal \cite{Wang14a} & {48.8} & 41.0 & 23.6 & 12.1 & 11.1 & 42.7 & 40.9 & {35.5} & \bf{11.1} & {36.6} & {18.4} & {35.3} & {34.8} & {51.3} & {17.2} & 17.4 & {26.8} & {32.8} & 35.1 & {45.6} & {30.9}\\
Wang \etal \cite{Wang14a}+context & \bf{48.9} & 42.3 & 26.1 & 11.3 & 11.9 & 41.3 & 40.9 & 34.7 & 10.8 & 34.7 & 18.8 & 34.4 & 35.4 & 52.7 & 19.1 & 17.4 & 35.9 & 33.3 & 34.8 & 46.5 & 31.6\\
\end{tabular}
}
\vspace{0.2em}
\caption{{\bf VOC 2007 test} detection average precision (\%). Comparison of our WSDDN on PASCAL VOC 2007 to the state-of-the-art in terms of AP.}
 \label{tab:voc2007ap}
\end{table*}

\begin{table*}[t!]
\centering
\renewcommand{\arraystretch}{1.2}
\renewcommand{\tabcolsep}{1.2mm}
\resizebox{\linewidth}{!}{
  \begin{tabular}{@{}L{3.5cm}|*{20}{x}|x@{}}
    method & aero      & bike      & bird      & boat      & bottle     & bus        & car        & cat        & chair      & cow        & table      & dog        & horse      & mbike      & persn     & plant      & sheep      & sofa       & train      & tv         & mean       \\
    \midrule
    WSDDN \bf{S} & 68.5 & 67.5 & 56.7 & 34.3 & 32.8 & 69.9 & 75.0 & 45.7 & 17.1 & 68.1 & 30.5 & 40.6 & 67.2 & 82.9 & 28.8 & 43.7 & 71.9 & 62.0 & 62.8 & 58.2 & 54.2\\
    WSDDN \bf{M} & 65.1 & 63.4 & 59.7 & {45.9} & 38.5 & 69.4 & 77.0 & 50.7 & 30.1 & \bf{68.8} & 34.0 & 37.3 & 61.0 & 82.9 & 25.1 & 42.9 & \bf{79.2} & 59.4 & \bf{68.2} & 64.1 & 56.1\\
    WSDDN \bf{L} & 65.1 & 58.8 & 58.5 & 33.1 & 39.8 & 68.3 & 60.2 & \bf{59.6} & \bf{34.8} & 64.5 & 30.5 & 43.0 & 56.8 & 82.4 & 25.5 & 41.6 & 61.5 & 55.9 & 65.9 & 63.7 & 53.5\\
    WSDDN Ensemble & 68.9 & \bf{68.7} & \bf{65.2} & 42.5 & \bf{40.6} & \bf{72.6} & {75.2} & {53.7} & {29.7} & {68.1} & {33.5} & {45.6} & {65.9} & \bf{86.1} & {27.5} & \bf{44.9} & {76.0} & \bf{62.4} & {66.3} & \bf{66.8} &\bf{58.0}\\
    \midrule
    Bilen \etal \cite{Bilen15}    & 66.4 & 59.3 & 42.7 & 20.4 & {21.3} & {63.4} & {74.3} & \bf{59.6} & 21.1 & {58.2} & 14.0 & 38.5 & 49.5 & 60.0 & 19.8 & {39.2} & 41.7 & 30.1 & 50.2 & 44.1 & 43.7\\
    Cinbis \etal \cite{Cinbis15} & 65.3 & 55.0 & 52.4 & \bf{48.3} & 18.2 & 66.4 & \bf{77.8} & 35.6 & 26.5 & 67.0 & \bf{46.9} & \bf{48.4} & \bf{70.5} & 69.1 & \bf{35.2} & 35.2 & 69.6 & 43.4 & 64.6 & 43.7 & 52.0 \\
    Wang \etal \cite{Wang14a}     & \bf{80.1} & {63.9} & {51.5} & 14.9 & 21.0 & 55.7 & 74.2 & 43.5 & {26.2} & 53.4 & 16.3 & 56.7 & {58.3} & {69.5} & 14.1 & 38.3 & {58.8} & {47.2} & 49.1 & {60.9} & {48.5}\\
  \end{tabular}
  }
  \vspace{0.2em}
  \caption{{\bf VOC 2007 trainval} correct localization (CorLoc \cite{Deselaers12}) on positive \textit{trainval} images (\%).}
  \label{tab:voc2007corloc}
\end{table*}

\begin{table*}[t!]
\centering
\renewcommand{\arraystretch}{1.2}
\renewcommand{\tabcolsep}{1.2mm}
\resizebox{\linewidth}{!}{
  \begin{tabular}{@{}L{3.5cm}|*{20}{x}|x@{}}
method & aero      & bike      & bird      & boat      & bottle     & bus        & car        & cat        & chair      & cow        & table      & dog        & horse      & mbike      & persn     & plant      & sheep      & sofa       & train      & tv         & mean       \\
\midrule
WSDDN \bf{S}   & 92.5 & 89.9 & 89.5 & 88.3 & 66.5 & 83.6 & 92.1 & 90.3 & 73.0 & 85.7 & 72.6 & 91.4 & 90.1 & 89.0 & 94.4 & 78.1 & 86.0 & 76.1 & 91.1 & 85.5 & 85.3\\
WSDDN \bf{M}   & 93.9 & 91.0 & 90.4 & 89.3 & 72.7 & 86.4 & 91.9 & 91.5 & 73.8 & 85.6 & 74.9 & 91.9 & 91.5 & 89.9 & 94.5 & 78.6 & 85.0 & 78.6 & 91.5 & 85.7 & 86.4\\ 
WSDDN \bf{L}   & 93.3 & \bf{93.9} & 91.6 & \bf{90.8} & \bf{82.5} & \bf{91.4} & \bf{92.9} & \bf{93.0} & 78.1 & \bf{90.5} & \bf{82.3} & \bf{95.4} & \bf{92.7} & 92.4 & 95.1 & \bf{83.4} & \bf{90.5} & 80.1 & 94.5 & \bf{89.6} & \bf{89.7}\\
WSDDN Ensemble & 95.0 & 92.6 & 91.2 & 90.4 & 79.0 & 89.2 & 92.8 & 92.4 & \bf{78.5} & \bf{90.5} & 80.4 & 95.1 & 91.6 & \bf{92.5} & 94.7 & 82.2 & 89.9 & \bf{80.3} & 93.1 & 89.1 & 89.0\\
\midrule
Oquab \etal \cite{Oquab14} &  88.5  &  81.5  &  87.9  &  82.0  &  47.5  &  75.5  &  90.1  &  87.2  &  61.6  &  75.7  &  67.3 &  85.5 &  83.5 &  80.0 &  \bf{95.6} &  60.8 &  76.8 &  58.0 &  90.4  &  77.9  &  77.7 \\
SPP \cite{He14} & -- &  -- & -- & -- & -- & -- & -- & -- & -- & -- & -- & -- & -- & -- & -- & -- & -- & -- & -- & -- & 82.4 \\
VGG-F \cite{Chatfield14} &  88.7 & 83.9 & 87.0 & 84.7 & 46.9 & 77.5 &  86.3 & 85.4 & 58.6 & 71.0 & 72.6 & 82.0 & 87.9 & 80.7 & 91.8 & 58.5 & 77.4 & 66.3 &  89.1 & 71.3 & 77.4\\
VGG-M-1024 \cite{Chatfield14} & 91.4 & 86.9 & 89.3 & 85.8 & 53.3 & 79.8 & 87.8 & 88.6 & 59.0 & 77.2 & 73.1 & 85.9 & 88.3 & 83.5 & 91.8 & 59.9 & 81.4 & 68.3 & 93.0 & 74.1 & 79.9\\
VGG-S \cite{Chatfield14} &  \bf{95.3} & {90.4} & \bf{92.5} & {89.6} & 54.4 & 81.9 & {91.5} & {91.9} & 64.1 & 76.3 & 74.9 & {89.7} & {92.2} & {86.9} & {95.2} & 60.7 & 82.9 & 68.0 & \bf{95.5} & 74.4 & 82.4\\
VGG-VD16 \cite{Simonyan15} & -- &  -- & -- & -- & -- & -- & -- & -- & -- & -- & -- & -- & -- & -- & -- & -- & -- & -- & -- & -- & 89.3 \\ 
\bottomrule
\end{tabular}
}
\vspace{0.05em}
\caption{{\bf VOC 2007 test} classification average precision (\%).}
 \label{tab:voc2007cls}
\end{table*}

\begin{table*}[t!]
\centering
\renewcommand{\arraystretch}{1.2}
\renewcommand{\tabcolsep}{1.2mm}
\resizebox{\linewidth}{!}{
  \begin{tabular}{@{}L{3.5cm}|*{20}{x}|x@{}}
    method & aero      & bike      & bird      & boat      & bottle     & bus        & car        & cat        & chair      & cow        & table      & dog        & horse      & mbike      & persn     & plant      & sheep      & sofa       & train      & tv         & mean       \\
    \midrule
    WSDDN Ensemble                      & \bf{57.4} & \bf{51.8} & \bf{41.2} & \bf{16.4} & \bf{22.8} & \bf{57.3} & \bf{41.8} & \bf{34.8} & \bf{13.1} & \bf{37.6} & 10.8 & \bf{37.0} & \bf{45.2} & \bf{64.9} & 14.1 & \bf{22.3} & \bf{33.8} & \bf{27.6} & \bf{49.1} & \bf{44.8} & \bf{36.2}\\
    \midrule
    Cinbis \etal \cite{Cinbis15} & 44.6 & 42.3 & 25.5 & 14.1 & 11.0 & 44.1 & 36.3 & 23.2 & 12.2 & 26.1 & \bf{14.0} & 29.2 & 36.0 & 54.3 & \bf{20.7} & 12.4 & 26.5 & 20.3 & 31.2 & 23.7 & 27.4\\
  \end{tabular}
  }
  \vspace{0.2em}
  \caption{{\bf VOC 2010 test} detection average precision (\%). \url{http://host.robots.ox.ac.uk:8080/anonymous/3QGEGM.html}}
  \label{tab:voc2010ap}
\end{table*}

\begin{table*}[t!]
\centering
\renewcommand{\arraystretch}{1.2}
\renewcommand{\tabcolsep}{1.2mm}
\resizebox{\linewidth}{!}{
  \begin{tabular}{@{}L{3.5cm}|*{20}{x}|x@{}}
    method & aero      & bike      & bird      & boat      & bottle     & bus        & car        & cat        & chair      & cow        & table      & dog        & horse      & mbike      & persn     & plant      & sheep      & sofa       & train      & tv         & mean       \\
    \midrule
    WSDDN Ensemble & \bf{77.4} & \bf{73.2} & \bf{61.9} & 39.6 & \bf{50.8} & \bf{84.4} & 67.5 & \bf{49.6} & 38.6 & \bf{73.4} & 30.4 & \bf{53.2} & 72.9 & \bf{84.1} & 30.3 & \bf{53.1} & \bf{76.6} & 48.5 & 61.6 & \bf{66.7} & \bf{59.7}\\
    \midrule
    Cinbis \etal \cite{Cinbis15} &  61.1 & 65.0 & 59.2 & \bf{44.3} & 28.3 & 80.6 & \bf{69.7} & 31.2 & \bf{42.8} & 73.3 & \bf{38.3} & 50.2 & \bf{74.9} & 70.9 & \bf{37.3} & 37.1 & 65.3 & \bf{55.3} & \bf{61.7} & 58.2 & 55.2\\
  \end{tabular}
  }
  \vspace{0.2em}
  \caption{{\bf VOC 2010 trainval} correct localization (CorLoc \cite{Deselaers12}) on positive \textit{trainval} images (\%).}
  \vspace{2em}
  \label{tab:voc2010corloc}
\end{table*}

\subsection{Detection results}\label{s:detres}

\paragraph{Baseline method.} First we design a single stream classification-detection network as an alternative baseline to WSDDN. Part of the construction is similar to WSDDN, as we replace \textit{pool5} layer of VGG-CNN-F model with an SPP. However, we do not branch off two streams, but simply append to the last fully connected layer ($\phi_\text{fc8c}$) the following loss layer \[
\frac{1}{nC}
\sum_{i=1}^n
\sum_{k=1}^C
\max\{0, 1 - y_{ki} \log \sum_{r=1}^{|\mathcal{R}|} \exp(x^{\mathcal{R}}_{cr})\}.
\]
The term $\log \sum_{r=1}^{|\mathcal{R}|} \exp(x^{\mathcal{R}}_{cr}$) is a soft approximation of the max operator $\max_r x_{cr}^\mathcal{R}$ and was found to yield better performance than using the max scoring region. This observation is also reported in \cite{Bilen14}. Note that the non-linearity is necessary as otherwise aggregating region-based scores would sum over the scores of a majority of regions that are uninformative. The loss function is once more a sum of $C$ binary hinge-losses, one for each class. This baseline obtains $21.6 \%$ mAP detection score on the PASCAL VOC test set, which is well below the state-of-the-art ($31.6 \%$ in~\cite{Wang14a}).

\paragraph{Pre-trained CNN architectures.} We evaluate our method with the models \textbf{S}, \textbf{M} and \textbf{L} and also report the results for the ensemble of these models by simply averaging their scores. Table~\ref{tab:voc2007base} shows that WSDDN with individual models \textbf{S} and \textbf{M} are already on par with the state-of-the-art method~\cite{Wang14a} and the ensemble outperforms the best previous score in the VOC 2007 dataset. Differently from supervised detection methods (\eg \cite{Girshick15}), detection performance of WSDDN does not improve with use of wider or deeper networks. In contrast, model \textbf{L} performs significantly worse than models \textbf{S} and \textbf{M} (see~\cref{tab:voc2007base}). This can be explained with the fact that model~\textbf{L} frequently focuses on parts of objects, instead of whole instances, and is still able to associate these parts with object categories due to its smaller convolution strides, higher resolution and deeper architecture.

\paragraph{Object proposals.} Next, we compare the detection performances with two popular object proposal methods, SSW~\cite{Sande11} and EB~\cite{Zitnick14}. While both the region proposals provides comparable quality region proposals, using box scores of EB (denoted as \emph{Box Sc} in~\cref{tab:voc2007base}) leads to a $~2\%$ improvement for models \textbf{S} and \textbf{M} and boosts the detection performance of model \textbf{L} $~5\%$.

\paragraph{Spatial regulariser.} We denote the setting where WSDDN is trained with the additional spatial regularisation term (denoted as \emph{Sp. Reg.} in \cref{tab:voc2007base}). Finally the introduction of the regularisation improves the detection performance $1$, $2$ and $4$ mAP points for models \textbf{S}, \textbf{M} and \textbf{L} respectively. The improvements show that larger network benefits more from introduction of the spatial invariance around high confidence regions.

\paragraph{Comparison with the state of the art.}
After evaluating the design decisions, we follow the best setting (last row in \cref{tab:voc2007base}) and compare WSDDN to the state of the art in weakly supervised detection literature in \cref{tab:voc2007ap} and \cref{tab:voc2007corloc} for the VOC 2007 dataset and in \cref{tab:voc2010ap} and \cref{tab:voc2010corloc} for the VOC 2010 dataset. The results show that our method already achieves overall significantly better performance than these alternatives with a single model and ensemble models further boost the performance. The majority of previous work \cite{Song14,Song14a,Bilen14,Wang14,Bilen15} use the Caffe reference CNN model \cite{Jia13}, which is comparable to model~\textbf{S} in this paper, as a black box to extract features over SSW proposals. In addition to CNN features, Cinbis \etal \cite{Cinbis15} use Fisher Vectors~\cite{Perronnin10a} and EB objectness measure of Zitnick and Dollar \cite{Zitnick14} as well. Differently from the previous work, WSDDN is based on a simple modification of the original CNN architecture fine-tuned on the target data using back-propagation.

\begin{figure*}[t!]
\begin{tabular}{*{6}{C{0.13\textwidth}}}
  \includegraphics[height=0.12\textwidth,width=0.15\textwidth]{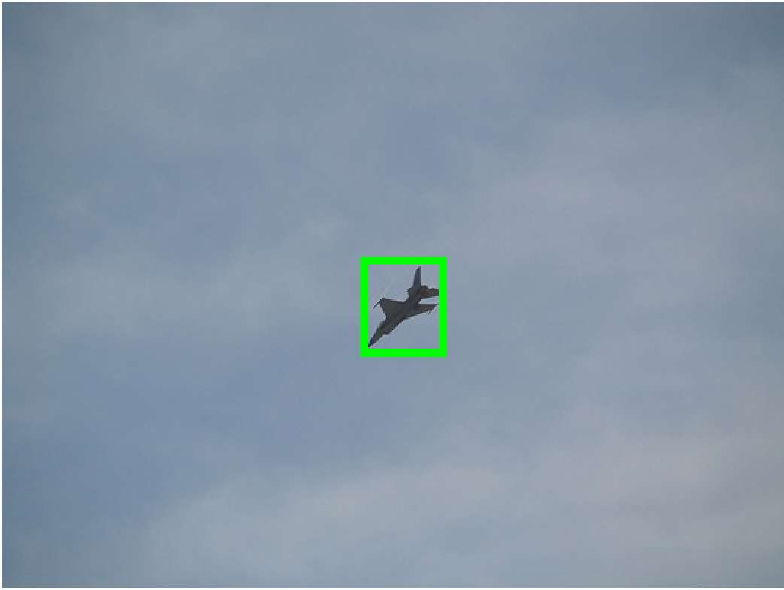}&
  \includegraphics[height=0.12\textwidth,width=0.15\textwidth]{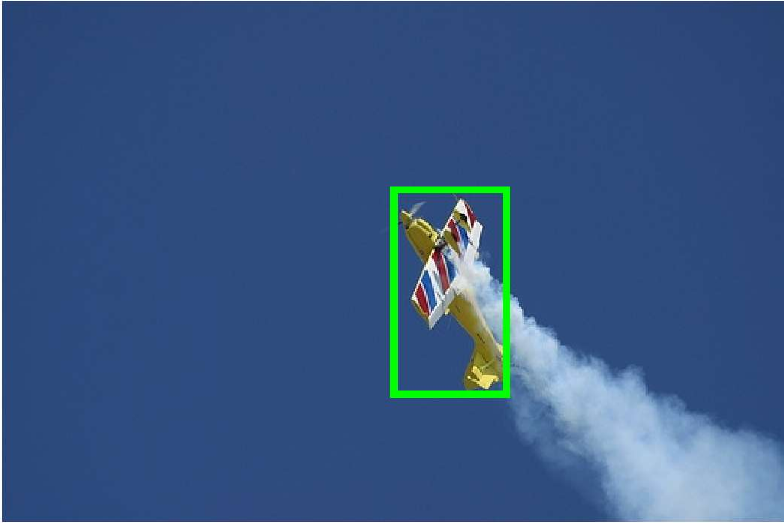}&
  \includegraphics[height=0.12\textwidth,width=0.15\textwidth]{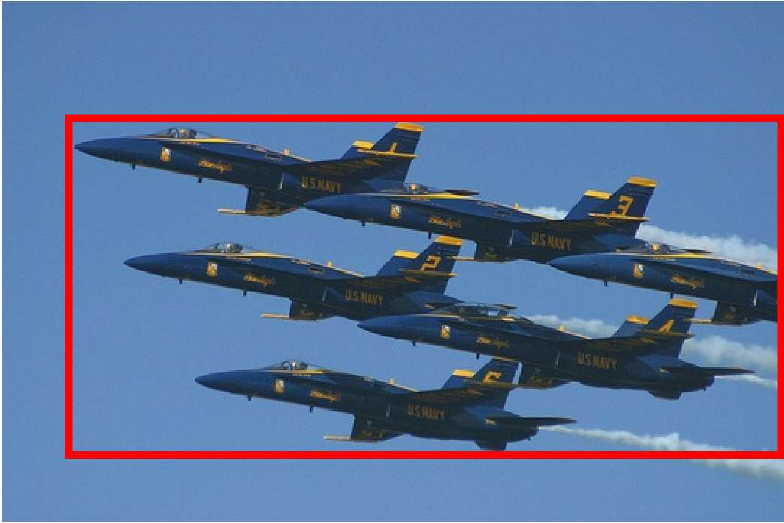}&
  \includegraphics[height=0.12\textwidth,width=0.15\textwidth]{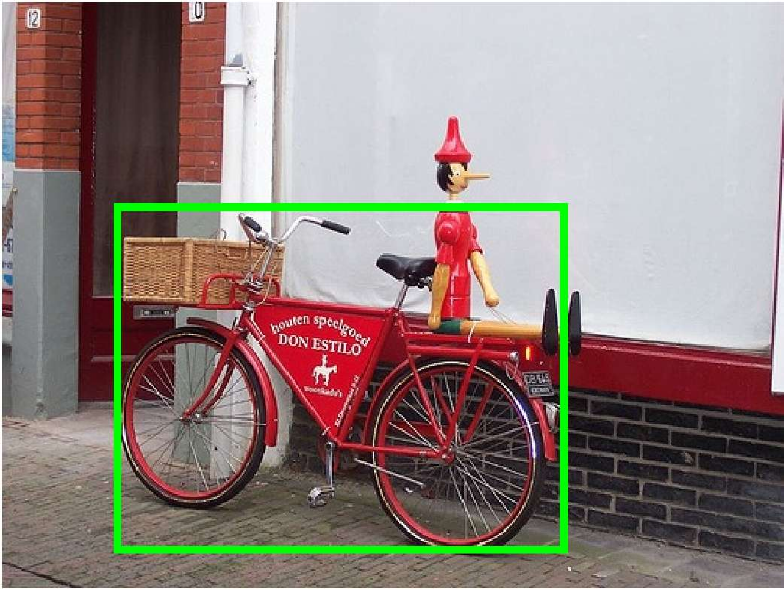}&
  \includegraphics[height=0.12\textwidth,width=0.15\textwidth]{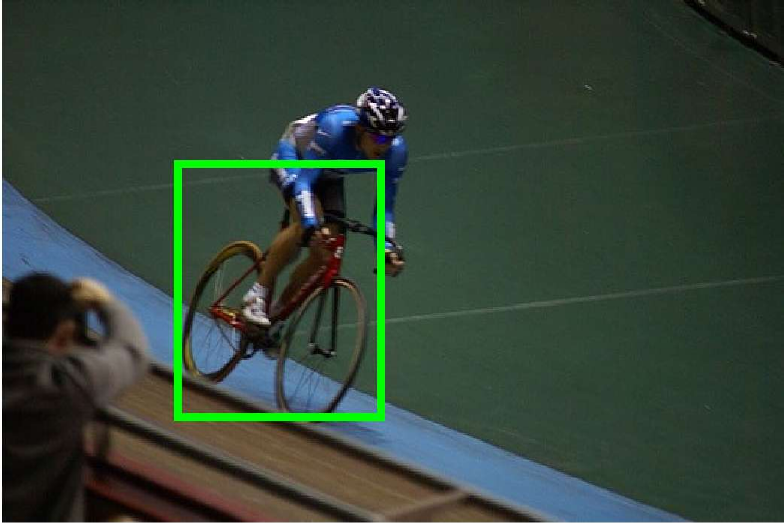}&
  \includegraphics[height=0.12\textwidth,width=0.15\textwidth]{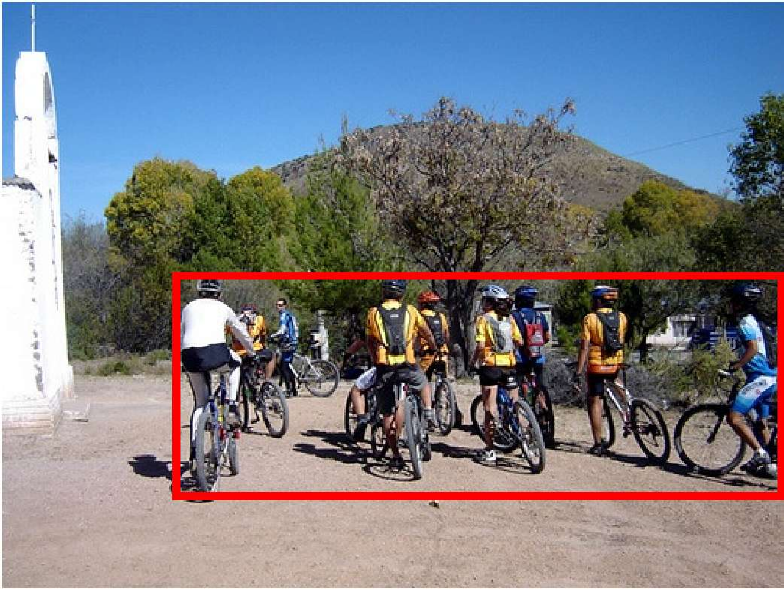}\\[-2ex]

  \includegraphics[height=0.12\textwidth,width=0.15\textwidth]{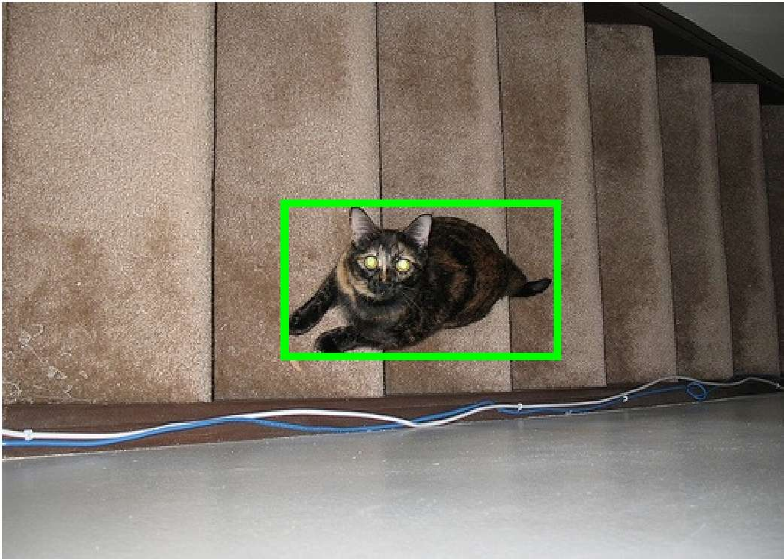}&
  \includegraphics[height=0.12\textwidth,width=0.15\textwidth]{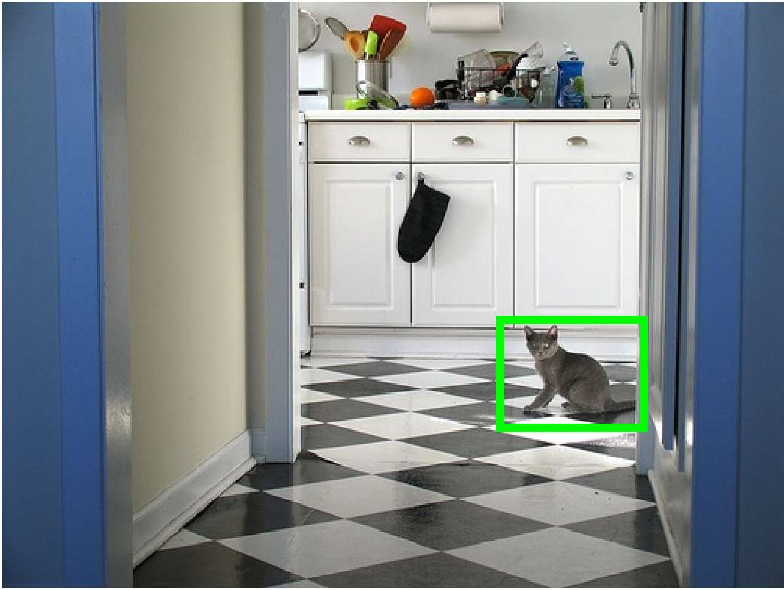}&
  \includegraphics[height=0.12\textwidth,width=0.15\textwidth]{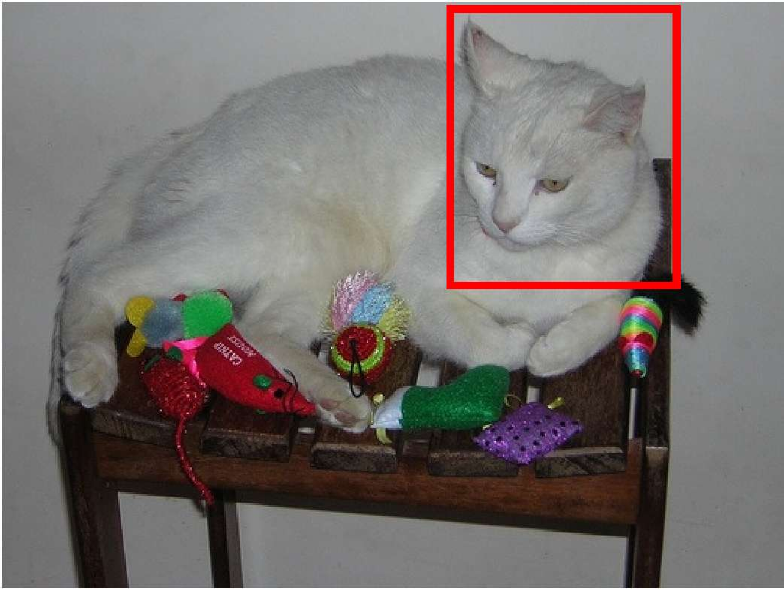}&
  \includegraphics[height=0.12\textwidth,width=0.15\textwidth]{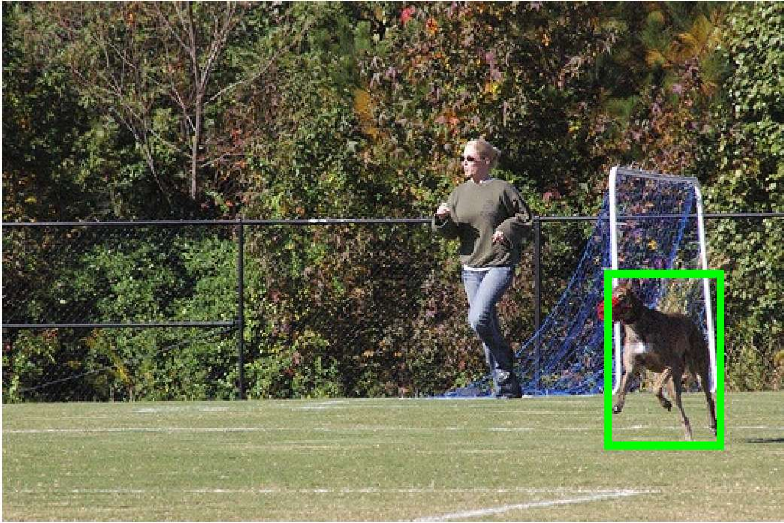}&
  \includegraphics[height=0.12\textwidth,width=0.15\textwidth]{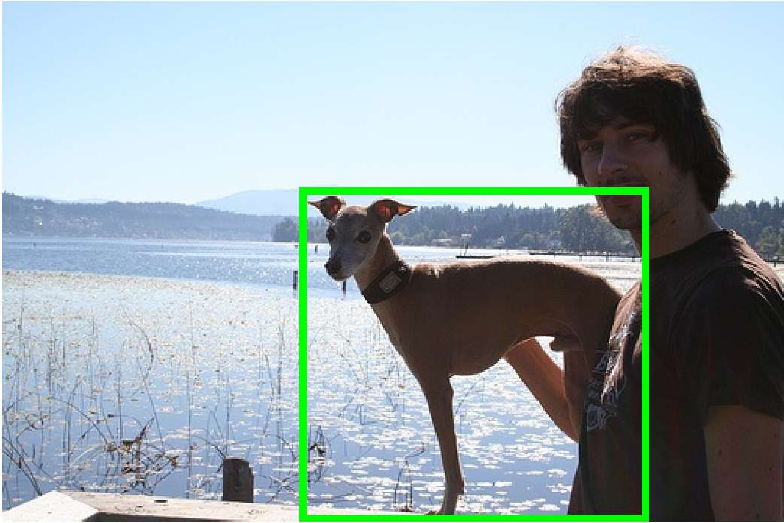}&
  \includegraphics[height=0.12\textwidth,width=0.15\textwidth]{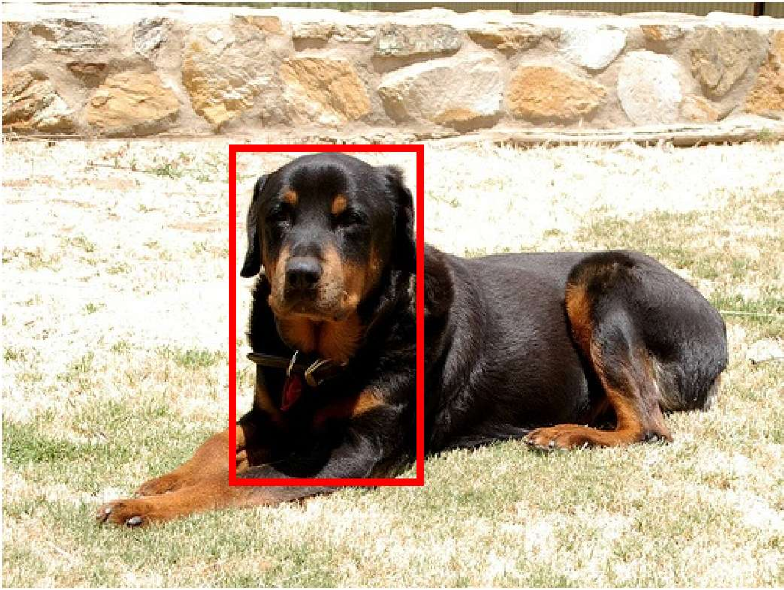}\\[-2ex]
  
  \includegraphics[height=0.12\textwidth,width=0.15\textwidth]{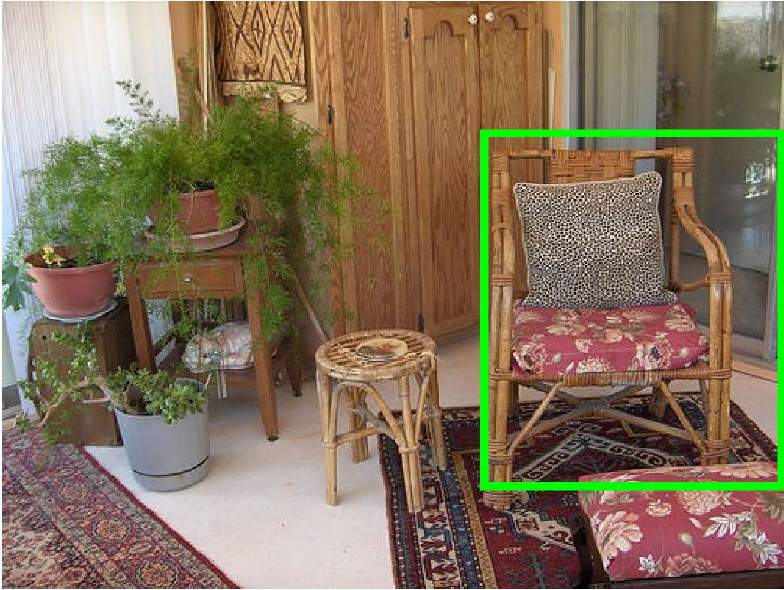}&
  \includegraphics[height=0.12\textwidth,width=0.15\textwidth]{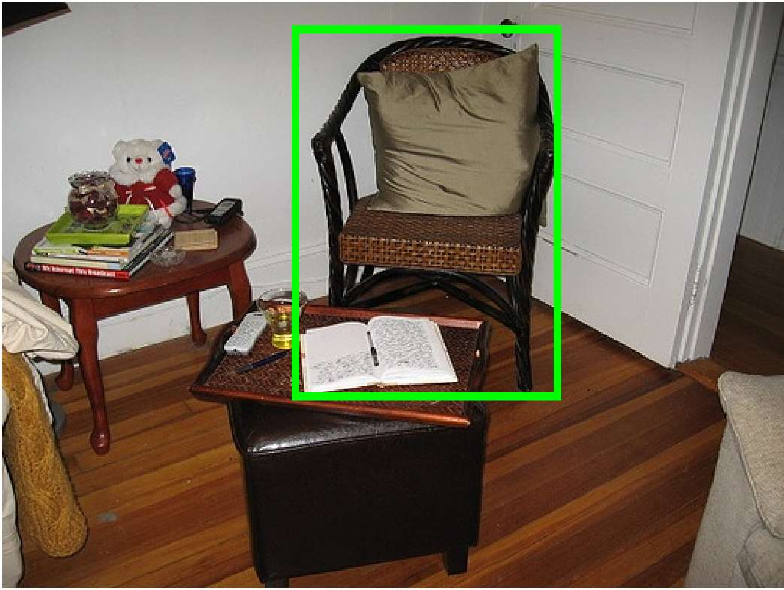}&
  \includegraphics[height=0.12\textwidth,width=0.15\textwidth]{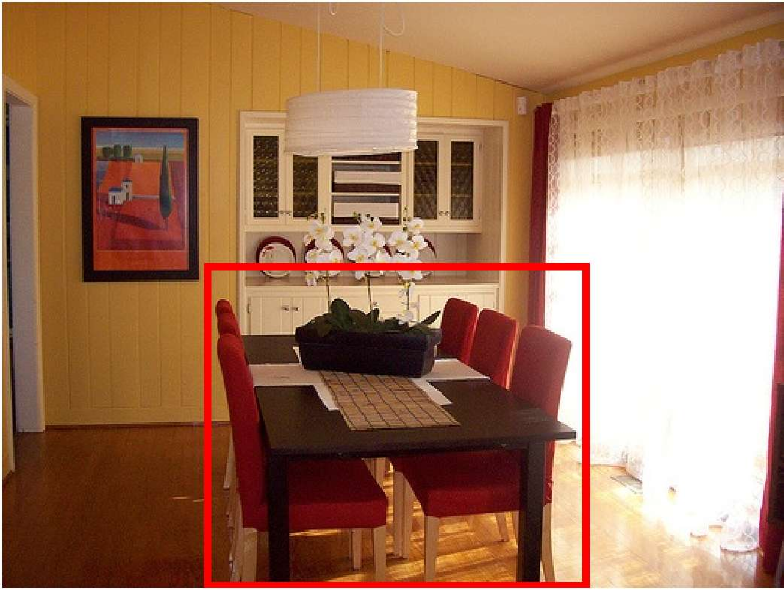}&
  \includegraphics[height=0.12\textwidth,width=0.15\textwidth]{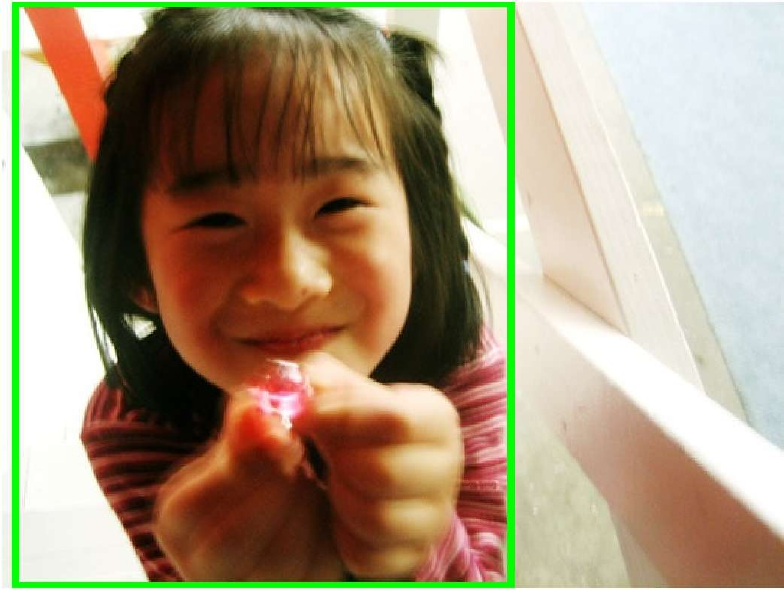}&
  \includegraphics[height=0.12\textwidth,width=0.15\textwidth]{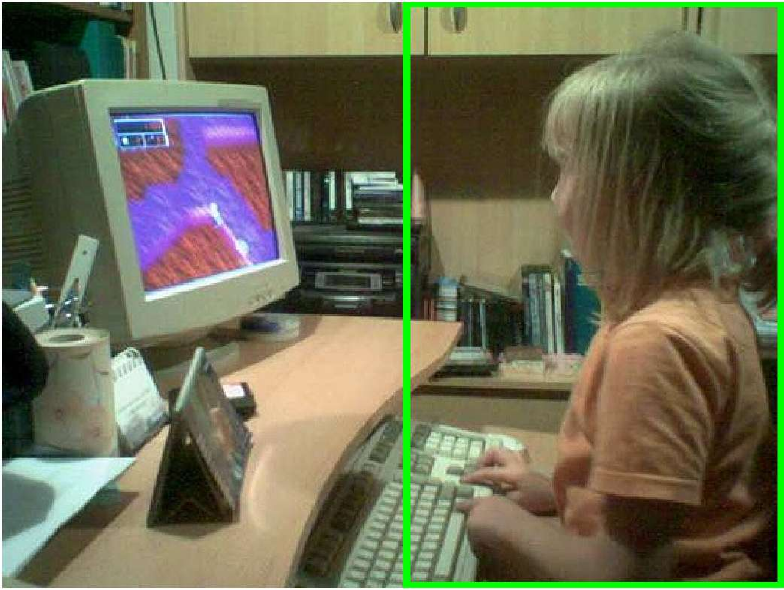}&
  \includegraphics[height=0.12\textwidth,width=0.15\textwidth]{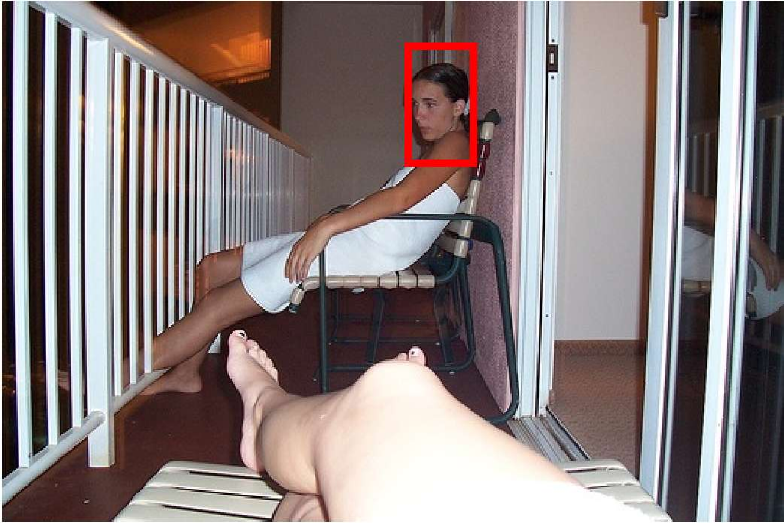}\\[-2ex]
  
  \includegraphics[height=0.12\textwidth,width=0.15\textwidth]{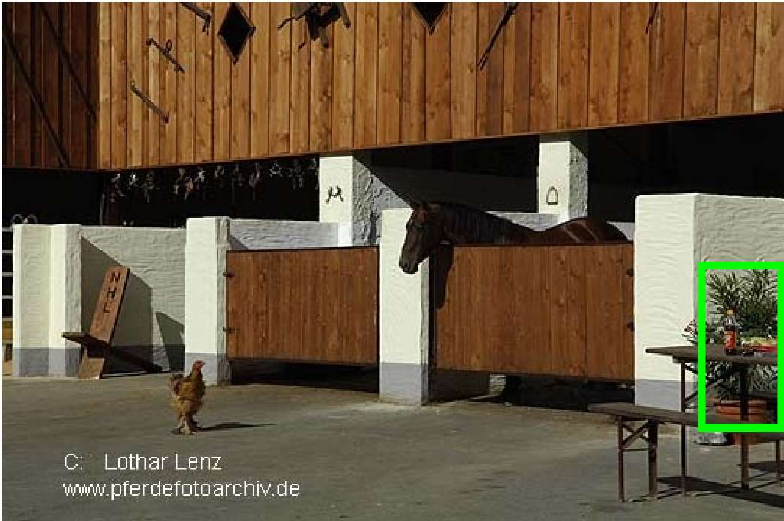}&
  \includegraphics[height=0.12\textwidth,width=0.15\textwidth]{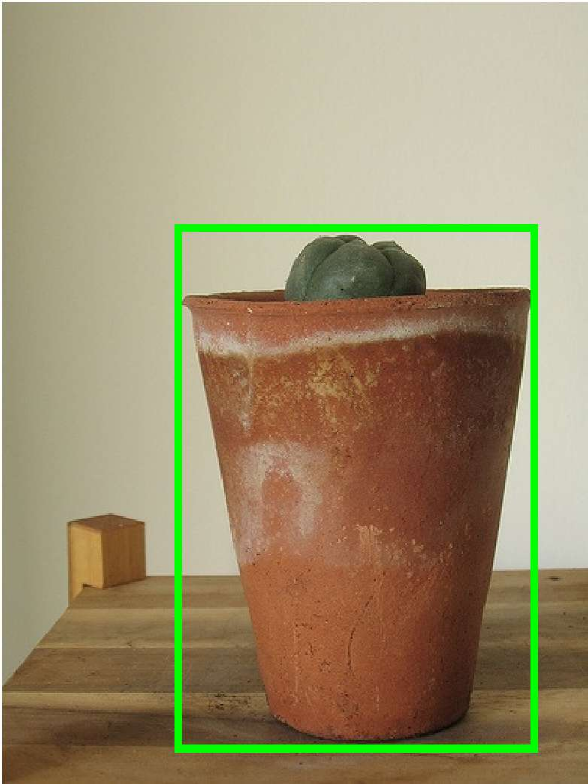}&
  \includegraphics[height=0.12\textwidth,width=0.15\textwidth]{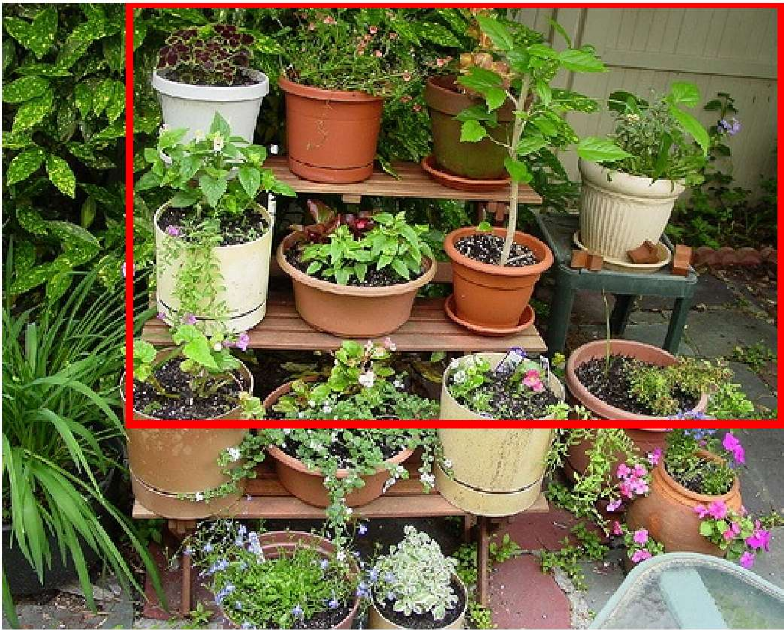}&
  \includegraphics[height=0.12\textwidth,width=0.15\textwidth]{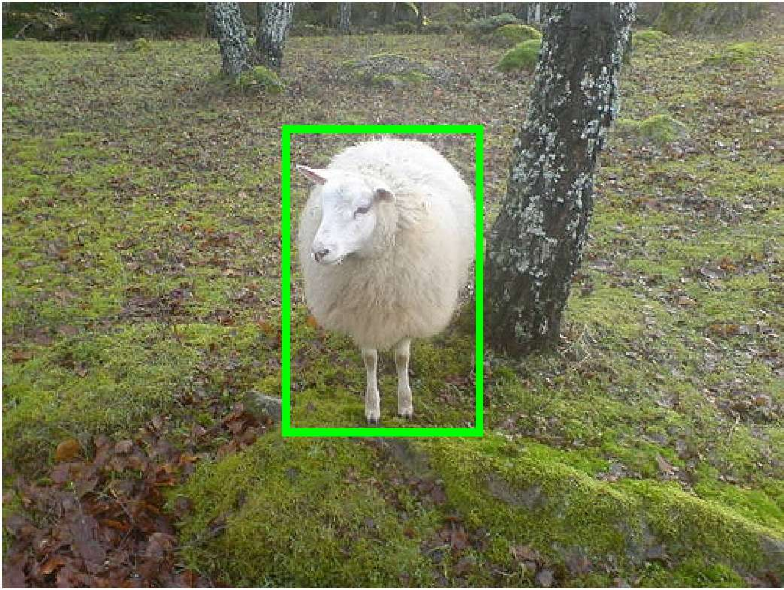}&
  \includegraphics[height=0.12\textwidth,width=0.15\textwidth]{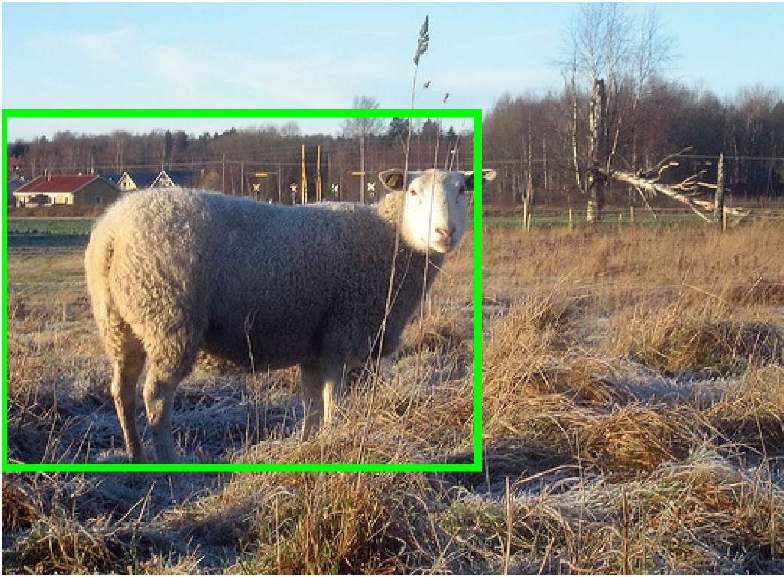}&
  \includegraphics[height=0.12\textwidth,width=0.15\textwidth]{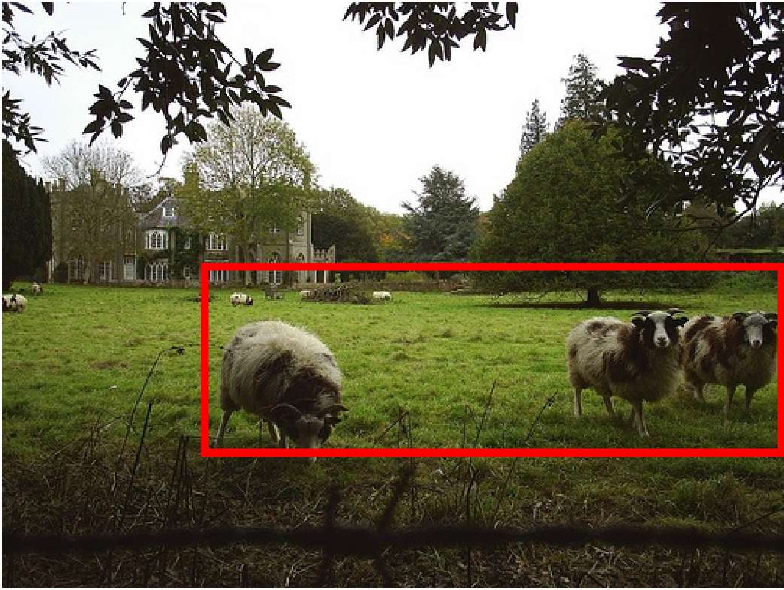}
  \end{tabular}
\caption{This figure depicts success (in green) and failure cases (in red) of our detector in randomly picked images. Majority of false detections contains two kinds of error: i) group multiple object instances with a single bounding box, ii) focus on (discriminative) parts (\eg ``faces'') rather than whole object. }
\label{fig:detexamples}
\end{figure*}

Next, we investigate the results in more detail. While our method significantly outperforms the alternatives in majority of categories, is not as strong in chair, person and pottedplant categories. Failure and success case are illustrated in \cref{fig:detexamples}. It can be noted that, by far, the most important failure modality for our system is that an object part (e.g. person face) is detected instead as the object as a whole. This can be explained by the fact that parts such as ``face'' are very often much more discriminative and with a less variable appearance than the rest of the object. Note that the root cause for this failure modality is that we, as many other authors, define objects as image regions that are most predictive for a given object class, and these may not include the object as a whole. Addressing this issue will therefore require incorporating additional cue in the model to try to learn the ``whole object''.

The output of our model could also be used as input to one of the existing methods for weakly-supervised detection that use a CNN as a black-box for feature extraction. Investigating this option is left to future work.

\subsection{Classification Results}
While WSDDN is primarily designed for weakly-supervised object detection, ultimately it is trained to perform image classification. Hence, it is interesting to evaluate its performance on this task as well. To this end, we use the PASCAL VOC 2007 benchmark and contrast it to standard fine-tuning techniques that are often used in combination with CNNs and show the results in \cref{tab:voc2010corloc}. These techniques have been thoroughly investigated in~\cite{Chatfield14,He14,Oquab14}. Chatfield \etal~\cite{Chatfield14}, in particular, analyse many variants of fine-tuning, including extensive data augmentation, on the PASCAL VOC. They experiment with three architectures, VGG-F, VGG-M, and VGG-S. While VGG-F is their fastest model, the other two networks are slower but more accurate. As explained in \ref{subsec:expsetup}, we initialise WSDDN \textbf{S} and \textbf{M} with the pre-trained VGG-F and VGG-M-1024 respectively and thus they should be considered as right baselines. WSDDN~\textbf{S} and \textbf{M} improves $~8$ and $~7$ points over VGG-F and VGG-M-1024 respectively.

We also compare WSDDN to the SPP-net \cite{He14} which uses the Overfeat-7 \cite{Sermanet13} with a 4-level spatial pyramid pooling layer $\{6\times 6, 3\times 3, 2\times 2, 1\times 1\}$ for supervised object detection. While they do not perform fine-tuning, they include a spatial pooling layer. Applied to image classification, their best performance on the PASCAL VOC 2007 is $82.4\%$. Finally we compare WSDDN~\textbf{L} to the competitive VGG-VD16~\cite{Simonyan15}. Interestingly, this method also exploits coarse local information by aggregating the activations of the last fully connected layer over multiple locations and scales. WSDDN~\textbf{L} outperforms this very competitive baseline with a margin of $0.4$ point.

\section{Conclusions}\label{s:conclusions}
In this paper, we have presented WSDDN, a simple modification of a pre-trained CNN for image classification that allows it to perform weakly supervised detection. It achieves significantly better performance than existing methods on weakly supervised detection, while requiring only fine-tuning on a target dataset using back-propagation, region proposals and image-level labels. Since it works on top of a SPP layer, it is also efficient at training and test time. WSDDN is also shown to perform better than traditional fine-tuning techniques to improve the performance of a pre-trained CNN on the problem of image classification.

We have identified the detection of object parts as a failure modality of the method, damaging its performance in selected object categories, and imputed that to the main criterion used to identify objects, namely the selection of highly-distinctive image regions. We are currently exploring complementary cues that would favour detecting complete objects instead.


{\small
\textbf{Acknowledgments:} This work acknowledges the support of the EPSRC EP/L024683/1, EPSRC Seebibyte EP/M013774/1 and the ERC Starting Grant IDIU. 
}

\clearpage
{
\small
\bibliographystyle{ieee}
\bibliography{shortstrings,vgg_local,vgg_other,mybib}
}

\end{document}